\documentclass[twocolumn,10pt]{article}
\usepackage{graphicx}
\usepackage{epstopdf}
\usepackage{algorithmic}
\usepackage{subfigure}
\usepackage{amsmath}
\usepackage{titling}
\usepackage{amssymb}
\usepackage{mathrsfs}
\usepackage[round]{natbib}
\usepackage[top=1in, bottom=1in, left=1in, right=1in]{geometry}
\usepackage{adjustbox}
\usepackage{array}

\title{\textbf{AC-BLSTM: Asymmetric Convolutional Bidirectional LSTM Networks for Text Classification}}

\author{Depeng~Liang$^1$\\
liangdp@mail2.sysu.edu.cn\\
\and
Yongdong~Zhang$^1$\thanks{
Corresponding author. E-mail: lnszyd@mail.sysu.edu.cn.}\\
lnszyd@mail.sysu.edu.cn\\}

\date{$^1$Guangdong Province Key Laboratory of Computational Science, School of Data and Computer Science, Sun Yat-sen University, Guang Zhou, China}

\begin{document}

\maketitle

\begin{abstract}
Recently deeplearning models have been shown
to be capable of making remarkable performance
in sentences and documents classification tasks. In this work, we propose
a novel framework called AC-BLSTM for modeling sentences and documents, which combines the asymmetric convolution neural network (ACNN) with the Bidirectional Long Short-Term Memory network (BLSTM). Experiment results demonstrate that our model achieves state-of-the-art results on five tasks, including sentiment analysis, question type classification, and subjectivity classification. In order to further improve the performance of AC-BLSTM, we propose a semi-supervised learning framework called G-AC-BLSTM for text classification by combining the generative model with AC-BLSTM.
\end{abstract}

\begin{figure*}[ht]
\begin{center}
\includegraphics[width=\textwidth]{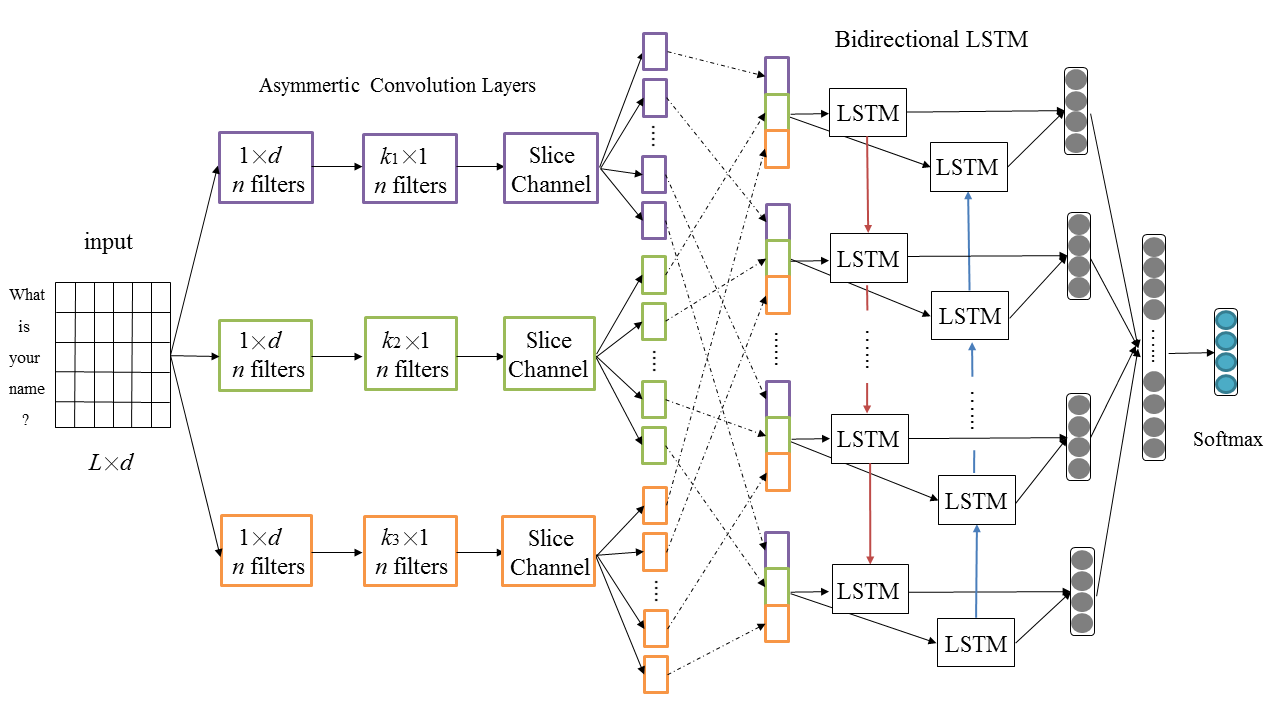}
\vspace{-4mm}
\caption{Illustration of the AC-BLSTM architecture. The input is represented as a matrix where each row is a d-dimensional word vector. Then the ACNN is applied to obtain the feature maps, we apply three parallel asymmetric convolution operation on the input in our model, where $k_{1}$, $k_{2}$ and $k_{3}$ stand for the length of the filter. And then the features with the same convolution window index from different convolution layer (different color) are concatenated to generate the input sequence of BLSTM. Finally all the hidden units of BLSTM are concatenated then apply a softmax layer to obtain the prediction output.}
\label{fig:overview}
\end{center}
\end{figure*}

\section{Introduction}
Deep neural models recently have achieved remarkable results in
computer vision \citep{AlexNet,GoogleNet,VGG,ResidualNet}, and a range of NLP tasks such as
sentiment classification \citep{Kim14,CLSTM,DCNN}, and question-answering \citep{E2EMemoryNet}. Convolutional neural networks (CNNs) and recurrent neural networks (RNNs) especially Long Short-term Memory Network (LSTM), are used wildly in natural language processing tasks. With increasing datas, these two methods can reach considerable performance by requiring only limited domain knowledge and easy to be finetuned to specific applications at the same time.

\par CNNs, which have the ability of capturing local correlations of spatial or temporal structures, have achieved excellent performance in computer vision and NLP tasks. And recently the emerge of some new techniques, such as Inception module \citep{InceptionV3}, Batchnorm \citep{BatchNorm} and Residual Network \citep{ResidualNet} have also made the performance even better. For sentence modeling, CNNs perform excellently in extracting n-gram features at different positions of a sentence through convolutional filters.

\par RNNs, with the ability of handling sequences of any length and capturing long-term dependencies, , have also achieved remarkable results in sentence or document modeling tasks. LSTMs \citep{LSTM1} were designed for better remembering and memory accesses, which can also avoid the problem of gradient exploding or vanishing in the standard RNN. Be capable of incorporating context on both sides of every position in the input sequence, BLSTMs introduced in \citep{BiLSTM1,BiLSTM4} have reported to achieve great performance in Handwriting Recognition \citep{BiLSTM3},
and Machine Translation \citep{BiLSTM2} tasks.

Generative adversarial networks (GANs) \citep{GAN} are a class of generative models for learning how to produce images. Basically, GANs consist of a generator G and a discriminator D, which are trained based on game theory. G maps a input noise vector to an output image, while D takes in an image then outputs a prediction whether the input image is a sample generated by G. Recently, applications of GANs have shown that they can generate promising results \citep{DCGAN,DPGAN}. Several recent papers have also extended GANs to the semi-supervised context \citep{Semi-GAN,Improve-GAN} by simply increasing the dimension of the classifier output from $K$ to $K + 1$, which the samples of the extra class are generated by G.

\par In this paper, We proposed an end-to-end architecture named AC-BLSTM by combining the ACNN with the BLSTM for sentences and documents modeling. In order to make the model deeper, instead of using the normal convolution, we apply the technique proposed in \citep{InceptionV3} which employs a $1 \times n$ convolution followed by a $n \times 1$ convolution by spatial factorizing the $n \times n$ convolution. And we use the pretrained word2vec vectors \citep{Word2vec} as the ACNN input, which were trained on 100 billion words of Google News to learn the higher-level representations of n-grams. The outputs of the ACNN are organized as the sequence window feature to feed into the multi-layer BLSTM. So our model does not rely on any other extra domain specific knowledge and complex preprocess, e.g. word segmentation, part of speech tagging and so on. We evaluate AC-BLSTM on sentence-level and document-level tasks including sentiment analysis, question type classification, and subjectivity classification. Experimental results demonstrate the effectiveness of our approach compared with other state-of-the-art methods. Further more, inspired by the ideas of extending GANs to the semi-supervised learning context by \citep{Semi-GAN,Improve-GAN}, we propose a semi-supervised learning framework for text classification which further improve the performance of AC-BLSTM.

\par The rest of the paper is organized as follows. Section 2 presents a brief review of related work. Section 3 discusses the architecture of our AC-BLSTM and our semi-supervised framework. Section 4 presents the experiments result with comparison analysis. Section 5 concludes the paper.

\section{Related Work}
Deep learning models have made remarkable progress in various NLP tasks recently. For example,
word embeddings \citep{Word2vec,Glove}, question answearing \citep{E2EMemoryNet}, sentiment analysis \citep{UPNN,NSCLA,SModel}, machine translation \citep{S2S} and so on. CNNs and RNNs are two wildly used architectures among these models. The success of deep learning models for NLP mostly relates to the progress in learning distributed word representations \citep{Word2vec,Glove}. In these mothods, instead of using one-hot vectors by indexing words into a vocabulary, each word is modeled as a low dimensional and dense vector which encodes both semantic and syntactic information of words.

\par Our model mostly relates to \citep{Kim14} which combines CNNs of different filter lengths and either static or fine-tuned word vectors, and \citep{CLSTM} which stacks CNN and LSTM in a unified architecture with static word vectors. It is known that in computer vision, the deeper network architecture usually possess the better performance. We consider NLP also has this property. In order to make our model deeper, we apply the idea of asymmetric convolution introduced in \citep{InceptionV3}, which can reduce the number of the parameters, and increase the representation ability of the model by adding more nonlinearity. Then we stack the multi-layer BLSTM, which is cable of analysing the future as well as the past of every position in the sequence, on top of the ACNN. The experiment results also demonstrate the effectiveness of our model.

\begin{figure}[ht]
\begin{center}
\includegraphics[width=0.45\textwidth]{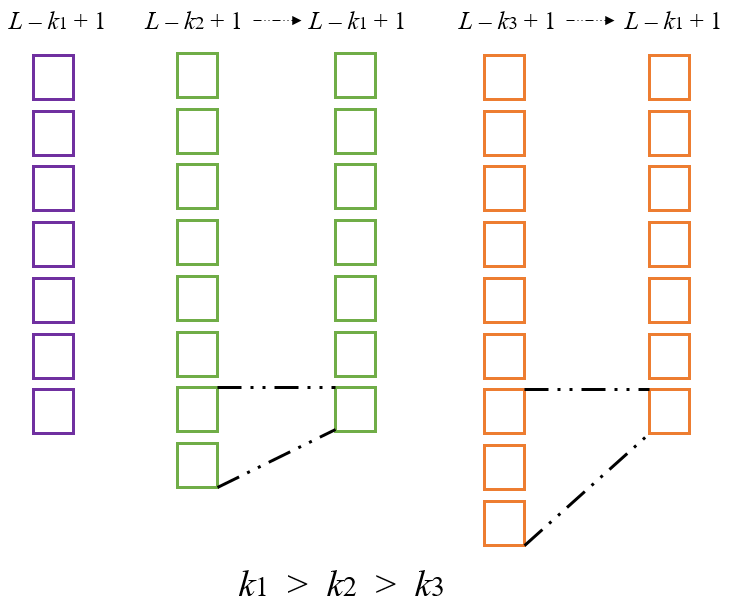}
\vspace{-4mm}
\caption{Illustration of how to deal with the incosistence output sequence length by compression.}
\label{fig:tricks}
\end{center}
\end{figure}

\section{AC-BLSTM Model}
In this section, we will introduce our AC-BLSTM architecture
in detail. We first describe the ACNN which takes the word vector represented matrix of the sentence as input and produces higher-level presentation of word features. Then we introduce the BLSTM which can incorporate context on both sides of every position in the input sequence.
Finally, we introduce the techniques to avoid overfitting in our model. An overall illustration of our architecture is shown in Figure ~\ref{fig:overview}.

\subsection{Asymmetric Convolution}
\par Let \textbf{x}$_{j} \in \mathbb{R}^{d}$ be the $d$-dimensional word vector corresponding to the $j$-th word in the sentence and $L$ be the maximum length of the sentence in the dataset. Then the sentence with length $L$ is represented as
\begin{equation}
\mathrm{\textbf{x}}_{1:L} = [\mathrm{\textbf{x}}_{1}, \mathrm{\textbf{x}}_{2},...,\mathrm{\textbf{x}}_{L}].
\end{equation}
For those sentences that are shorter than $L$, we simply pad them with space.
\par In general, let $k_{i}$ in which $i \in \{1,2,3\}$ be the length of convolution filter. Then instead of employing the ${k_i \times d}$ convolution operation described in \citep{Kim14,CLSTM}, we apply the asymmetric convolution operation inspired by \citep{InceptionV3} to the input matrix which factorize the ${k_i \times d}$ convolution into $1 \times d$ convolution followed by a $k_i \times 1$ convolution. And in experiments, we found that employ this technique can imporve the performance.
The following part of this subsection describe how we define the asymmetric convolution layer.

\par First, the convolution operation corresponding to the $1 \times d$ convolution with
filter \textbf{w}$_i^1 \in \mathbb{R}^{d}$ is applied to each word \textbf{x}$_j$ in the sentence and generates corresponding feature \textbf{m}$_j^i$
\begin{equation}
\mathrm{\textbf{m}}_j^i = f(\mathrm{\textbf{w}}_i^1 \circ \mathrm{\textbf{x}}_j+b).
\end{equation}
where $\circ$ is element-wise multiplication, $b$ is a bias term and $f$ is a non-linear function
such as the sigmoid, hyperbolic tangent, etc. In our case, we choose ReLU \citep{Relu}
as the nonlinear function. Then we get the feature map \textbf{m}$^i \in \mathbb{R}^L$

\begin{equation}
\mathrm{\textbf{m}}^i = [\mathrm{\textbf{m}}_1^i, \mathrm{\textbf{m}}_2^i,..., \mathrm{\textbf{m}}_L^i].
\end{equation}
\par After that, the second convolution operation of the asymmetric convolution layer corresponding to the $k_i \times 1$ convolution with filter \textbf{w}$_i^2 \in \mathbb{R}^{k_i}$ is applied to
a window of $k_i$ features in the feature map \textbf{m}$^i$ to produce the new feature \textbf{c}$_j^i$ and the feature map \textbf{c}$^i$
\begin{equation}
\mathrm{\textbf{c}}_j^i = f(\mathrm{\textbf{w}}_i^2 \circ \mathrm{\textbf{m}}_{j:j+k_i-1}^i+b).
\end{equation}

\begin{equation}
\mathrm{\textbf{c}}^i = [\mathrm{\textbf{c}}_1^i, \mathrm{\textbf{c}}_2^i,..., \mathrm{\textbf{c}}_{L-k_i+1}^i].
\end{equation}
with \textbf{c}$^i \in \mathbb{R}^{L-k_i+1}$. Where $\circ$, $b$ and $f$ are the same as described above.
\par As shown in Figure ~\ref{fig:overview}, we simultaneously apply three asymmetric convolution layers to the input matrix, which all have the same number of filters denoted as $n$. Thus the output of the asymmetric convolution layer has $n$ feature maps. To generate the input sequence of the BLSTM, for each output sequence of the second convolution operation in the aysmmetric convolution layer, we slice the feature maps by channel then obtained sequence of $L-k_i+1$ new features \textbf{c}$_t^i \in \mathbb{R}^{n}$ where $t \in \{1,2,...,L-k_i+1\}$. Then we concatanate \textbf{c}$_t^1$, \textbf{c}$_t^2$ and \textbf{c}$_t^3$ to get the input feature for each time step
\begin{equation}
\mathrm{\hat{\textbf{c}}}_t = [\mathrm{\textbf{c}}_t^1,\mathrm{\textbf{c}}_t^2, \mathrm{\textbf{c}}_t^3].
\end{equation}
where $\mathrm{\hat{\textbf{c}}}_t \in \mathbb{R}^{3n}$ for $t \in \{1,2,...,L-\hat{k}+1\}$ and $\hat{k} = \max\limits_i{k_i}$. In general, those \textbf{c}$_t^i$ where $k_i < \hat{k}$ and $t > L-\hat{k}+1$ must be dropped in order to maintain the same sequence length, which will cause the loss of some information. In our model, instead of simply cutting the sequence, we use a simple trick to obtain the same sequence length without losing the useful information as shown in Figure ~\ref{fig:tricks}. For each output sequence $\mathrm{\textbf{c}}_t^i$ obtained from the second convolution operation with filter length $k_i$, we take those \textbf{c}$_t^i$ where $t >= L-\hat{k}+1$ then apply a fullyconnected layer to get a new feature, which has the same dimension of \textbf{c}$_t^i$, to replace the ($L-\hat{k}$+1)-th feature in the origin sequence.

\subsection{Bidirectional Long Short-Term Memory Network}
First introduced in \citep{LSTM1} and shown as a successful model recently, LSTM is a RNN architecture specifically designed to bridge long time delays between relevant input and target events, making it suitable for problems where long range context is required, such as handwriting recognition, machine translation and so on.

\par For many sequence processing tasks, it is useful to analyze the future as well as the past of a given point in the series. Whereas standard RNNs make use of previous context only, BLSTM \citep{BiLSTM1} is explicitly designed for learning long-term dependencies of a given point on both side, which has also been shown to outperform other neural network architectures in framewise phoneme recognition \citep{BiLSTM4}.
\par Therefore we choose BLSTM on top of the ACNN to learn such dependencies given the sequence of higher-level features. And single layer BLSTM can extend to multi-layer BLSTM easily. Finally, we concatenate all hidden state of all the time step of BLSTM, or concatenate the last layer of all the time step hidden state of multi-layer BLSTM, to obtain final representation of the text and we add a softmax layer on top of the model for classification.

\subsection{Semi-supervised Framework}
Our semi-supervised text classification framewrok is inspired by works \citep{Semi-GAN,Improve-GAN}. We assume the original classifier classify a sample into one of $K$ possible classes. So we can do semi-supervised learning by simply adding samples from a generative network G to our dataset and labeling them to an extra class $y = K + 1$. And correspondingly the dimension of our classifier output increases from $K$ to $K + 1$. The configuration of our generator network G is inspired by the architecture proposed in \citep{DCGAN}. And we modify the architecture to make it suitable to the text classification tasks. Table~\ref{table:gan} shows the configuration of each layer in the generator G. Lets assume the training batch size is $m$ and the percentage of the generated samples among a batch training samples is $p_g$. At each iteration of the training process, we first generate $m \times p_g$ samples from the generator G then we draw $m - m \times p_g$ samples from the
real dataset. We then perform gradient descent on the AC-BLSTM and generative net G and finally update the parameters of both nets.

\subsection{Regularization}
For model regularization, we employ two commonly used techniques to prevent overfitting during training: dropout \citep{Dropout} and batch normalization \citep{BatchNorm}. In our model, we apply dropout to the input feature of the BLSTM, and the output of BLSTM before the softmax layer. And we apply batch normalization to outputs of each convolution operation just before the relu activation.
During training, after we get the gradients of the AC-BLSTM network, we first calculate the $L2$ $norm$ of all gradients and sum together to get $sum\_norm$. Then we compare the $sum\_norm$ to 0.5. If the $sum\_norm$ is greater than 0.5, we let all the gradients multiply with $0.5 / sum\_norm$, else just use the original gradients to update the weights.

\begin{table*}[ht]
\centering
\begin{adjustbox}{width=\textwidth}
\begin{tabular}{l|c|c|c|c|c|c}
	\hline
	\hline
		\textbf{Layer} 	            & Input &  Filter & stride & Output	\\
	\hline
		Fc and Reshape              & 100  &  - &  -  &  $h \times w \times c_g$  \\
		Deconv                      &  $h \times w \times c_g$   &   $4 \times 4$  &   $2 \times 2$   &   $2h \times 2w \times \frac{1}{2}c_g$   \\
		Deconv                		& $2h \times 2w \times \frac{1}{2}c_g$   &   $4 \times 4$  &   $2 \times 2$   &   $4h \times 4w \times \frac{1}{4}c_g$   \\
		Deconv                		&  $4h \times 4w \times \frac{1}{4}c_g$   &   $4 \times 4$  &   $2 \times 2$   &   $L \times d \times 1$   \\
	\hline
	\hline
\end{tabular}
\end{adjustbox}
\caption{Configuration of the generative network. $h$ and $w$ means the height and width of the output feature map of the first layer and set to
$\lfloor L / 4 \rfloor$ and $\lfloor d / 4 \rfloor$ initially. $c_g$ means the output channel number which needs to be tuned.}
\label{table:gan}
\end{table*}

\begin{table*}[ht]
\centering
\begin{adjustbox}{width=\textwidth}
\begin{tabular}{l|c|c|c|c|c|c}
	\hline
	\hline
		\textbf{Hyper-parameters} 	& TREC &  MR  & SST-1 & SST-2 & SUBJ &  YELP13	\\
	\hline
		convolution filters num     & 100  &  100 &  100  &  300  & 100  &  100 	\\
		lstm memory dimention       & 100  &  100 &  100  &  300  & 100  &  100 	\\
		lstm layer                  &  1   &   4  &   4   &   4   &  4   &   4		\\
		dropout before softmax		&  0.5 &  0.5 &  0.5  &  0.6  & 0.6  &  0.5		\\
		
	\hline
	\hline
\end{tabular}
\end{adjustbox}
\caption{Hyper-parameters setting of AC-BLSTM across datasets.}
\label{table:hyper-parameter}
\end{table*}

\begin{table*}[ht]
\centering
\begin{adjustbox}{width=\textwidth}
\begin{tabular}{l|c|c|c|c}
	\hline
	\hline
		\textbf{Hyper-parameters} 	&  MR  & SST-1 & SST-2 & SUBJ 	\\
	\hline
		$c_g$                  &   100  &   100   &   100   &  100   \\
		$p_g$		           &   20\% &   10\%  &   10\%  &  5\% 	 \\
		
	\hline
	\hline
\end{tabular}
\end{adjustbox}
\caption{Hyper-parameters setting of Generative net across datasets.}
\label{table:g-hyper-parameter}
\end{table*}

\begin{table*}[ht]
\centering
\begin{adjustbox}{width=\textwidth}
\begin{tabular}{l|c|c|c|c|c|c}
	\hline
	\hline
		\textbf{Model} 				                       & TREC &  MR  & SST-1 & SST-2 & SUBJ & YELP13	\\
	\hline
		CNN-non-static\citep{Kim14}                        & 93.6 & 81.5 & 48.0  & 87.2  & 93.4 & - \\
		CNN-multichannel\citep{Kim14}                      & 92.2 & 81.1 & 47.4  & 88.1  & 93.2 & -  \\	
		C-LSTM\citep{CLSTM} 		                       & 94.6 &  -   & 49.2  & 87.8  &  -   & -  \\
		Molding-CNN\citep{MoldingCNN}                      &  -   &  -   & 51.2  & 88.6  &  -   & -  \\
		UPNN(no UP)\citep{UPNN}		                       &  -   &  -   &  -    & 	-    &  -   & 57.7  \\
		DSCNN\citep{DSCNN}		   		                   & 95.4 & 81.5 & 49.7  & 89.1  & 93.2 & -  \\
		DSCNN-Pretrain\citep{DSCNN}		                   & 95.6 & 82.2 & 50.6  & 88.7  & 93.9 & -  \\
		MG-CNN(w2v+Syn+Glv)\citep{MGCNN}                   & 94.68&  -   & 48.01 & 87.63 & 94.11& -  \\
		MGNC-CNN(w2v+Glv)\citep{MGCNN} 	                   & 94.40&  -   & 48.53 & 88.35 & 93.93& -  \\
		MGNC-CNN(w2v+Syn+Glv)\citep{MGCNN}                 & 95.52&  -   & 48.65 & 88.30 & 94.09& -  \\
		NSC+LA\citep{NSCLA}				                   & -    &  -   &  -    &  -    &  -   & 63.1  \\
		SequenceModel(no UP)\citep{SModel}                 & -    &  -   &  -    &  -    &  -   & 62.4  \\
        TreeBiGRU(with attention)\citep{TreeBiGRU}         & -    &  -   & 52.4  & 89.5  &  -   & -  \\
        TopCNN\textsubscript{word}\citep{TopCNN}           & 92.5 & 81.7 &  -    &  -    & 93.4 & -  \\
        TopCNN\textsubscript{sen}\citep{TopCNN}            & 92.0 & 81.3 &  -    &  -    & 93.4 & -  \\
        TopCNN\textsubscript{word\&sen}\citep{TopCNN}      & 93.6 & 82.3 &  -    &  -    & 94.3 & -  \\
        TopCNN\textsubscript{ens}\citep{TopCNN}            & 94.1 & 83.0 &  -    &  -    & \textbf{95.0} & -  \\
	\hline	
		AC-BLSTM(our model)                                &\textbf{95.8}&83.1&52.9&91.1&94.2&\textbf{63.6}\\
        G-AC-BLSTM(our model)                              & - &\textbf{83.7}&\textbf{53.2}&\textbf{91.5}&94.3&-\\
	\hline
	\hline
\end{tabular}
\end{adjustbox}
\caption{Experiment results of our AC-BLSTM and G-AC-BLSTM model compared with other methods. Performance is measured in accuracy. \textbf{CNN-non-static, CNN-multichannel}: Convolutional neural network with fine-tuned word vectors and multi-channels \citep{Kim14}.
\textbf{C-LSTM}: Combining CNN and LSTM to model sentences \citep{CLSTM}.
\textbf{Molding-CNN}: A feature mapping operation based on tensor products on stacked vectors \citep{MoldingCNN}. \textbf{UPNN(no UP)}: User product neural network without using user and product information \citep{UPNN}. \textbf{DSCNN, DSCNN-Pretrain}: Dependency sensitive convolutional neural networks and with pretraind sequence autoencoders \citep{DSCNN}. \textbf{MG-CNN(w2v+Syn+Glv), MGNC-CNN(w2v+Glv), MGNC-CNN(w2v+Syn+Glv)}: Multi-group
norm constraint CNN with w2v:word2vec, Glv:GloVe \citep{Glove} and Syn: Syntactic embedding \citep{MGCNN}. \textbf{NSC+LA}: Neural sentiment classification model with local semantic attention \citep{NSCLA}. \textbf{SequenceModel(no UP)}: A sequence modeling-based
neural network without using user and product information \citep{SModel}. \textbf{TreeBiGRU}: A tree-structured attention recursive neural networks that incorporates a bidirectional approach with gated memory units for sentence classification \citep{TreeBiGRU}. \textbf{TopCNN\textsubscript{word}, TopCNN\textsubscript{sen}, TopCNN\textsubscript{word\&sen}, TopCNN\textsubscript{ens}}: Topic-aware convolutional neural network for sentence classification. \textbf{TopCNN\textsubscript{word}} means to use the word-topic probability information to enrich the word embeddings. \textbf{TopCNN\textsubscript{sen}} means to use the sentence-topic probability information to enrich the representation output of the pooling layer. \textbf{TopCNN\textsubscript{word\&sen}} means to use both word-topic and sentence-topic probability information. \textbf{TopCNN\textsubscript{ens}} means an ensemble model of the three variants of TopCNN models by averaging the class probability scores generated by the three models together \citep{TopCNN}.}
\label{table:results}
\end{table*}

\section{Experiments}
\subsection{Datasets}
We evaluate our model on various benchmarks. Stanford Sentiment Treebank (SST) is a popular sentiment classification dataset introduced by \citep{SSTDataset}. The sentences are labeled in a fine-grained way (SST-1): {very negative, negative, neutral, positive, very positive}. The dataset has been split into 8,544 training, 1,101 validation, and 2,210 testing sentences. By  removing the neutral sentences, SST can also be used for binary classification (SST-2), which has been split into 6,920 training, 872 validation, and 1,821 testing. Since the data is provided in the format of sub-sentences, we train the model on both phrases and sentences but only test on the sentences as in several previous works \citep{SSTDataset, DCNN}.

\par Movie Review Data (MR) proposed by \citep{MRDataset} is another dataset for sentiment analysis of movie reviews. The dataset consists of 5,331 positive and 5,331 negative reviews, mostly in one sentence. We follow the practice of using 10-fold cross validation to report the result.

\par Furthermore, we apply AC-BLSTM on the subjectivity classification dataset (SUBJ) released by \citep{SubjDataset}. The dataset contains 5,000 subjective sentences and 5,000 objective sentences. We also follow the practice of using 10-fold cross validation to report the result.

\par We also benchmark our system on question type classification task (TREC) \citep{TRECDataset}, where sentences are questions in the following 6 classes: {abbreviation, human, entity, description, location, numeric}. The entire dataset consists of 5,452 training examples and 500 testing examples.

\par For document-level dataset, we use the sentiment classification dataset Yelp 2013 (YELP13) with user and product information, which is built by \citep{UPNN}. The dataset has been split into 62,522 training, 7,773 validation, and 8,671 testing documents. But in the experiment, we neglect the user and product information to make it consistent with the above experiment settings.

\subsection{Training and Implementation Details}
We implement our model based on Mxnet \citep{MXNet} - a C++ library, which is
a deep learning framework designed for both efficiency and flexibility. In order to benefit from the efficiency of parallel computation of the tensors, we train our model on a Nvidia GTX 1070 GPU. Training is done through stochastic gradient descent over shuffled mini-batches with the optimizer RMSprop \citep{RmsProp}. For all experiments, we simultaneously apply three asymmetric convolution operation with the second filter length $k_i$ of 2, 3, 4 to the input, set the dropout rate to 0.5 before feeding the feature into BLSTM, and set the initial learning rate to 0.0001. But there are some hyper-parameters that are not the same for all datasets, which are listed in table~\ref{table:hyper-parameter}. We conduct experiments on 3 datasets (MR, SST and SUBJ) to verify the effectiveness our semi-supervised framework.
And the setting of $p_g$ and $c_g$ for different datasets are listed in table~\ref{table:g-hyper-parameter}.

\subsection{Word Vector Initialization}
We use the publicly available word2vec vectors that were trained on 100 billion words from Google News. The vectors have dimensionality of 300 and were trained using the continuous bag-of-words
architecture \citep{Word2vec}. Words not present in the set of pre-trained words are initialized
from the uniform distribution [-0.25, 0.25]. We fix the word vectors and learn only the other parameters of the model during training.

\subsection{Results and Discussion}
We used standard train/test splits for those datasets that had them. Otherwise, we performed 10-fold cross validation. We repeated each experiment 10 times and report the mean accuracy. Results of our models against other methods are listed in table~\ref{table:results}. To the best of our knowledge, AC-BLSTM achieves the best results on five tasks.

\par Compared to methods \citep{Kim14} and \citep{CLSTM}, which inspired our model mostly,
AC-BLSTM can achieve better performance which show that deeper model actually has better performance.
By just employing the word2vec vectors, our model can achieve better results than \citep{MGCNN} which combines multiple word embedding methods such as word2vec\citep{Word2vec}, glove \citep{Glove} and Syntactic embedding. And the AC-BLSTM performs better when trained with the semi-supervised framework, which proves
the success of combining the generative net with AC-BLSTM.

\par The experiment results show that the number of the convolution filter and the lstm memory dimension should keep the same for our model. Also the configuration of hyper-parameters: number of the convolution filter, the lstm memory dimension and the lstm layer are quiet stable across datasets. If the task is simple, e.g. TREC, we just set number of convolution filter to 100, lstm memory dimension to 100 and lstm layer to 1. And as the task becomes complicated, we simply increase the lstm layer from 1 to 4. The SST-2 is a special case, we find that if we set the number of convolution filter and lstm memory dimension to 300 can get better result. And the dropout rate before softmax need to be tuned.

\section{Conclusions}
In this paper we have proposed AC-BLSTM: a novel framework that combines asymmetric convolutional neural network with bidirectional long short-term memory network. The asymmetric convolutional layers are able to learn phrase-level features. Then output sequences of such higher level representations are fed into the BLSTM to learn long-term dependencies of a given point on both side. To the best of our knowledge, the AC-BLSTM model achieves top performance on standard sentiment classification, question classification and document categorization tasks. And then we proposed a semi-supervised framework for text classification which further improve the performance of AC-BLSTM. In future work, we plan to explore the combination of multiple word embeddings which are described in \citep{MGCNN}.

\begingroup
    \setlength{\bibsep}{2pt}
    \bibliography{References}

\begin{thebibliography}{39}
\providecommand{\natexlab}[1]{#1}
\providecommand{\url}[1]{\texttt{#1}}
\expandafter\ifx\csname urlstyle\endcsname\relax
  \providecommand{\doi}[1]{doi: #1}\else
  \providecommand{\doi}{doi: \begingroup \urlstyle{rm}\Url}\fi

\bibitem[Chen et~al.(2016{\natexlab{a}})Chen, Sun, Tu, Lin, and Liu]{NSCLA}
Huimin Chen, Maosong Sun, Cunchao Tu, Yankai Lin, and Zhiyuan Liu.
\newblock Neural sentiment classification with user and product attention.
\newblock In \emph{EMNLP}, 2016{\natexlab{a}}.

\bibitem[Chen et~al.(2016{\natexlab{b}})Chen, Xu, He, Xia, and Wang]{SModel}
Tao Chen, Ruifeng Xu, Yulan He, Yunqing Xia, and Xuan Wang.
\newblock Learning user and product distributed representations using a
  sequence model for sentiment analysis.
\newblock \emph{IEEE Computational Intelligence Magazine}, 11\penalty0
  (3):\penalty0 34--44, 2016{\natexlab{b}}.

\bibitem[Chen et~al.(2015)Chen, Li, Li, Lin, Wang, Wang, Xiao, Xu, Zhang, and
  Zhang]{MXNet}
Tianqi Chen, Mu~Li, Yutian Li, Min Lin, Naiyan Wang, Minjie Wang, Tianjun Xiao,
  Bing Xu, Chiyuan Zhang, and Zheng Zhang.
\newblock Mxnet: A flexible and efficient machine learning library for
  heterogeneous distributed systems.
\newblock \emph{CoRR}, abs/1512.01274, 2015.

\bibitem[Denton et~al.(2015)Denton, Chintala, Szlam, and Fergus]{DPGAN}
Emily~L. Denton, Soumith Chintala, Arthur Szlam, and Rob Fergus.
\newblock Deep generative image models using a laplacian pyramid of adversarial
  networks.
\newblock In \emph{Advances in Neural Information Processing Systems 28: Annual
  Conference on Neural Information Processing Systems 2015, December 7-12,
  2015, Montreal, Quebec, Canada}, pages 1486--1494, 2015.

\bibitem[Goodfellow et~al.(2014)Goodfellow, Pouget{-}Abadie, Mirza, Xu,
  Warde{-}Farley, Ozair, Courville, and Bengio]{GAN}
Ian~J. Goodfellow, Jean Pouget{-}Abadie, Mehdi Mirza, Bing Xu, David
  Warde{-}Farley, Sherjil Ozair, Aaron~C. Courville, and Yoshua Bengio.
\newblock Generative adversarial nets.
\newblock In \emph{Advances in Neural Information Processing Systems 27: Annual
  Conference on Neural Information Processing Systems 2014, December 8-13 2014,
  Montreal, Quebec, Canada}, pages 2672--2680, 2014.

\bibitem[Graves and Schmidhuber(2005)]{BiLSTM4}
Alex Graves and J{\"{u}}rgen Schmidhuber.
\newblock Framewise phoneme classification with bidirectional {LSTM} and other
  neural network architectures.
\newblock \emph{Neural Networks}, 18\penalty0 (5-6):\penalty0 602--610, 2005.

\bibitem[Graves et~al.(2005)Graves, Fern{\'{a}}ndez, and Schmidhuber]{BiLSTM1}
Alex Graves, Santiago Fern{\'{a}}ndez, and J{\"{u}}rgen Schmidhuber.
\newblock Bidirectional {LSTM} networks for improved phoneme classification and
  recognition.
\newblock In \emph{ICANN}, pages 799--804, 2005.

\bibitem[He et~al.(2015)He, Zhang, Ren, and Sun]{ResidualNet}
Kaiming He, Xiangyu Zhang, Shaoqing Ren, and Jian Sun.
\newblock Deep residual learning for image recognition.
\newblock In \emph{arXiv prepring arXiv:1506.01497}, 2015.

\bibitem[Hochreiter and Schmidhuber(1997)]{LSTM1}
Sepp Hochreiter and J{\"{u}}rgen Schmidhuber.
\newblock Long short-term memory.
\newblock \emph{Neural Computation}, 9\penalty0 (8):\penalty0 1735--1780, 1997.

\bibitem[Ioffe and Szegedy(2015)]{BatchNorm}
Sergey Ioffe and Christian Szegedy.
\newblock Batch normalization: Accelerating deep network training by reducing
  internal covariate shift.
\newblock In \emph{ICML}, pages 448--456, 2015.

\bibitem[Kalchbrenner et~al.(2014)Kalchbrenner, Grefenstette, and
  Blunsom]{DCNN}
Nal Kalchbrenner, Edward Grefenstette, and Phil Blunsom.
\newblock A convolutional neural network for modelling sentences.
\newblock In \emph{Proceedings of the 52nd Annual Meeting of the Association
  for Computational Linguistics, {ACL} 2014, June 22-27, 2014, Baltimore, MD,
  USA, Volume 1: Long Papers}, pages 655--665, 2014.

\bibitem[Kim(2014)]{Kim14}
Yoon Kim.
\newblock Convolutional neural networks for sentence classification.
\newblock In \emph{EMNLP}, pages 1746--1751, 2014.

\bibitem[Kokkinos and Potamianos(2017)]{TreeBiGRU}
Filippos Kokkinos and Alexandros Potamianos.
\newblock Structural attention neural networks for improved sentiment analysis.
\newblock \emph{CoRR}, abs/1701.01811, 2017.

\bibitem[Krizhevsky et~al.(2012)Krizhevsky, Sutskever, and Hinton]{AlexNet}
Alex Krizhevsky, Ilya Sutskever, and Geoffrey~E. Hinton.
\newblock Imagenet classification with deep convolutional neural networks.
\newblock In \emph{Advances in Neural Information Processing Systems 25: 26th
  Annual Conference on Neural Information Processing Systems 2012. Proceedings
  of a meeting held December 3-6, 2012, Lake Tahoe, Nevada, United States.},
  pages 1106--1114, 2012.

\bibitem[Lei et~al.(2015)Lei, Barzilay, and Jaakkola]{MoldingCNN}
Tao Lei, Regina Barzilay, and Tommi Jaakkola.
\newblock Molding cnns for text: non-linear, non-consecutive convolutions.
\newblock In \emph{Proceedings of the 2015 Conference on Empirical Methods in
  Natural Language Processing}, September 2015.

\bibitem[Li and Roth(2002)]{TRECDataset}
Xin Li and Dan Roth.
\newblock Learning question classifiers.
\newblock In \emph{Proceedings of the 19th international conference on
  Computational linguistics-Volume 1}, pages 1--7, 2002.

\bibitem[Liwicki et~al.(2007)Liwicki, Graves, Bunke, and Schmidhuber]{BiLSTM3}
Marcus Liwicki, Alex Graves, Horst Bunke, and J{\"u}rgen Schmidhuber.
\newblock A novel approach to on-line handwriting recognition based on
  bidirectional long short-term memory networks.
\newblock In \emph{Proc. 9th Int. Conf. on Document Analysis and Recognition},
  volume~1, pages 367--371, 2007.

\bibitem[Mikolov et~al.(2013)Mikolov, Sutskever, Chen, Corrado, and
  Dean]{Word2vec}
Tomas Mikolov, Ilya Sutskever, Kai Chen, Gregory~S. Corrado, and Jeffrey Dean.
\newblock Distributed representations of words and phrases and their
  compositionality.
\newblock In \emph{Advances in Neural Information Processing Systems 26: 27th
  Annual Conference on Neural Information Processing Systems 2013. Proceedings
  of a meeting held December 5-8, 2013, Lake Tahoe, Nevada, United States.},
  pages 3111--3119, 2013.

\bibitem[Nair and Hinton(2010)]{Relu}
Vinod Nair and Geoffrey~E. Hinton.
\newblock Rectified linear units improve restricted boltzmann machines.
\newblock In \emph{ICML}, pages 807--814, 2010.

\bibitem[Odena(2016)]{Semi-GAN}
Augustus Odena.
\newblock Semi-supervised learning with generative adversarial networks.
\newblock \emph{arXiv preprint arXiv:1606.01583}, 2016.

\bibitem[Pang and Lee(2004)]{SubjDataset}
Bo~Pang and Lillian Lee.
\newblock A sentimental education: Sentiment analysis using subjectivity
  summarization based on minimum cuts.
\newblock In \emph{Proceedings of the 42nd annual meeting on Association for
  Computational Linguistics}, page 271, 2004.

\bibitem[Pang and Lee(2005)]{MRDataset}
Bo~Pang and Lillian Lee.
\newblock Seeing stars: Exploiting class relationships for sentiment
  categorization with respect to rating scales.
\newblock In \emph{ACL}, 2005.

\bibitem[Pennington et~al.(2014)Pennington, Socher, and Manning]{Glove}
Jeffrey Pennington, Richard Socher, and Christopher~D. Manning.
\newblock Glove: Global vectors for word representation.
\newblock In \emph{Proceedings of the 2014 Conference on Empirical Methods in
  Natural Language Processing, {EMNLP} 2014, October 25-29, 2014, Doha, Qatar,
  {A} meeting of SIGDAT, a Special Interest Group of the {ACL}}, pages
  1532--1543, 2014.

\bibitem[Peris and Casacuberta(2015)]{BiLSTM2}
Alvaro Peris and Francisco Casacuberta.
\newblock A bidirectional recurrent neural language model for machine
  translation.
\newblock \emph{Procesamiento del Lenguaje Natural}, 55:\penalty0 109--116,
  2015.

\bibitem[Radford et~al.(2015)Radford, Metz, and Chintala]{DCGAN}
Alec Radford, Luke Metz, and Soumith Chintala.
\newblock Unsupervised representation learning with deep convolutional
  generative adversarial networks.
\newblock \emph{CoRR}, abs/1511.06434, 2015.

\bibitem[Salimans et~al.(2016)Salimans, Goodfellow, Zaremba, Cheung, Radford,
  and Chen]{Improve-GAN}
Tim Salimans, Ian~J. Goodfellow, Wojciech Zaremba, Vicki Cheung, Alec Radford,
  and Xi~Chen.
\newblock Improved techniques for training gans.
\newblock In \emph{Advances in Neural Information Processing Systems 29: Annual
  Conference on Neural Information Processing Systems 2016, December 5-10,
  2016, Barcelona, Spain}, pages 2226--2234, 2016.

\bibitem[Simonyan and Zisserman(2014)]{VGG}
Karen Simonyan and Andrew Zisserman.
\newblock Very deep convolutional networks for large-scale image recognition.
\newblock \emph{CoRR}, abs/1409.1556, 2014.

\bibitem[Socher et~al.(2013)Socher, Perelygin, Wu, Chuang, Manning, Ng, and
  Potts]{SSTDataset}
Richard Socher, Alex Perelygin, Jean~Y Wu, Jason Chuang, Christopher~D Manning,
  Andrew~Y Ng, and Christopher Potts.
\newblock Recursive deep models for semantic compositionality over a sentiment
  treebank.
\newblock In \emph{EMNLP}, volume 1631, page 1642, 2013.

\bibitem[Srivastava et~al.(2014)Srivastava, Hinton, Krizhevsky, Sutskever, and
  Salakhutdinov]{Dropout}
Nitish Srivastava, Geoffrey~E. Hinton, Alex Krizhevsky, Ilya Sutskever, and
  Ruslan Salakhutdinov.
\newblock Dropout: a simple way to prevent neural networks from overfitting.
\newblock \emph{Journal of Machine Learning Research}, 15\penalty0
  (1):\penalty0 1929--1958, 2014.

\bibitem[Sukhbaatar et~al.(2015)Sukhbaatar, Szlam, Weston, and
  Fergus]{E2EMemoryNet}
Sainbayar Sukhbaatar, Arthur Szlam, Jason Weston, and Rob Fergus.
\newblock End-to-end memory networks.
\newblock In \emph{Advances in Neural Information Processing Systems 28: Annual
  Conference on Neural Information Processing Systems 2015, December 7-12,
  2015, Montreal, Quebec, Canada}, pages 2440--2448, 2015.

\bibitem[Sutskever et~al.(2014)Sutskever, Vinyals, Le, Sutskever, Vinyals, and
  Le]{S2S}
Ilya Sutskever, Oriol Vinyals, Quoc~V. Le, Ilya Sutskever, Oriol Vinyals, and
  Quoc~V. Le.
\newblock Sequence to sequence learning with neural networks.
\newblock \emph{Advances in Neural Information Processing Systems}, 4:\penalty0
  3104--3112, 2014.

\bibitem[Szegedy et~al.(2015{\natexlab{a}})Szegedy, Liu, Jia, Sermanet, Reed,
  Anguelov, Erhan, Vanhoucke, and Rabinovich]{GoogleNet}
Christian Szegedy, Wei Liu, Yangqing Jia, Pierre Sermanet, Scott~E. Reed,
  Dragomir Anguelov, Dumitru Erhan, Vincent Vanhoucke, and Andrew Rabinovich.
\newblock Going deeper with convolutions.
\newblock In \emph{{IEEE} Conference on Computer Vision and Pattern
  Recognition, {CVPR} 2015, Boston, MA, USA, June 7-12, 2015}, pages 1--9,
  2015{\natexlab{a}}.

\bibitem[Szegedy et~al.(2015{\natexlab{b}})Szegedy, Vanhoucke, Ioffe, Shlens,
  and Wojna]{InceptionV3}
Christian Szegedy, Vincent Vanhoucke, Sergey Ioffe, Jonathon Shlens, and
  Zbigniew Wojna.
\newblock Rethinking the inception architecture for computer vision.
\newblock \emph{CoRR}, abs/1512.00567, 2015{\natexlab{b}}.

\bibitem[Tang et~al.(2015)Tang, Qin, and Liu]{UPNN}
Duyu Tang, Bing Qin, and Ting Liu.
\newblock Learning semantic representations of users and products for document
  level sentiment classification.
\newblock In \emph{ACL}, pages 1014--1023, 2015.

\bibitem[Tieleman and Hinton(2012)]{RmsProp}
T.~Tieleman and G.~Hinton.
\newblock {Lecture 6.5---RmsProp: Divide the gradient by a running average of
  its recent magnitude}.
\newblock COURSERA: Neural Networks for Machine Learning, 2012.

\bibitem[Zhang et~al.(2016{\natexlab{a}})Zhang, Lee, and Radev]{DSCNN}
Rui Zhang, Honglak Lee, and Dragomir Radev.
\newblock Dependency sensitive convolutional neural networks for modeling
  sentences and documents.
\newblock In \emph{Proceedings of NAACL-HLT}, pages 1512--1521,
  2016{\natexlab{a}}.

\bibitem[Zhang et~al.(2016{\natexlab{b}})Zhang, Roller, and Wallace]{MGCNN}
Ye~Zhang, Stephen Roller, and Byron~C. Wallace.
\newblock Mgnc-cnn: A simple approach to exploiting multiple word embeddings
  for sentence classification.
\newblock In \emph{Proceedings of the 2016 Conference of the North American
  Chapter of the Association for Computational Linguistics: Human Language
  Technologies}, 2016{\natexlab{b}}.

\bibitem[Zhao and Mao(2017)]{TopCNN}
Rui Zhao and Kezhi Mao.
\newblock Topic-aware deep compositional models for sentence classification.
\newblock \emph{{IEEE/ACM} Trans. Audio, Speech {\&} Language Processing},
  25\penalty0 (2):\penalty0 248--260, 2017.

\bibitem[Zhou et~al.(2015)Zhou, Sun, Liu, and Lau]{CLSTM}
Chunting Zhou, Chonglin Sun, Zhiyuan Liu, and Francis C.~M. Lau.
\newblock A {C-LSTM} neural network for text classification.
\newblock \emph{CoRR}, abs/1511.08630, 2015.

\end{thebibliography}
\endgroup

\end{document}